\pdfoutput=1
\documentclass[11pt]{article}

\usepackage{acl}

\usepackage{times}
\usepackage{latexsym}
\usepackage{amsmath}
\usepackage{amssymb} 
\usepackage{booktabs} % For better table lines
\usepackage{tabularx} % For adjustable width columns
\usepackage{multirow}
\usepackage{adjustbox}
\usepackage{listings}
\usepackage[skins,most]{tcolorbox}
\tcbuselibrary{skins}
\usepackage{verbatim}
\usepackage{enumitem}
\usepackage{CJK}

\definecolor{lightred}{HTML}{111111}   %边框
\definecolor{custombg}{HTML}{F3F3F3}      % 背景颜色
\definecolor{red}{HTML}{FAD9D5}

\usepackage[T1]{fontenc}
\usepackage[utf8]{inputenc}

\usepackage{microtype}

\usepackage{inconsolata}

\usepackage{graphicx}
\usepackage{subcaption}

\title{Proactive Guidance of Multi-Turn Conversation in Industrial Search}

\author{
  \textbf{Xiaoyu Li, Xiao Li, Li Gao\thanks{Corresponding author.}, Yiding Liu, Xiaoyang Wang} \\
  \textbf{Shuaiqiang Wang, Junfeng Wang, Dawei Yin} \\
  \text{Baidu Inc., Beijing, China} \\
  \texttt{demo.xyli@icloud.com, \{emilyxiao0512, gaoli.sinh, liuyidingyd\}@gmail.com,} \\
  \texttt{\{wangxiaoyang06, wangshuaiqiang, wangjunfeng\}@baidu.com, yindawei@acm.org}
}

\begin{document}
\maketitle
%\begin{abstract}
%Proactive guidance in multi-turn conversation requires dynamic adaptation to evolving user goals while maintaining industrial-grade latency. This paper presents a novel two-phase framework combining Goal Adaptive Supervised Fine-Tuning (G-SFT) and Click Orinented Reinforcement Learning (C-RL) to address these challenges. In the G-SFT phase, we propose a Goal Adaptation Agent (GAA) that detects shifts in user intent through iterative dialogue analysis and generates concise, goal-aligned summaries to preserve contextual relevance. Coupled with Scalable Knowledge Transfer—which distills world knowledge from large language models (LLMs) into a compact model—this phase achieves XX. The C-RL phase introduces a Generate-Rank paradigm that leverages user click behavior for preference alignment. Using a preference-augmented model and a click prediction module with Group Sampling, we systematically construct high-quality preference pairs to optimize engagement. Our framework demonstrates significant improvements over baseline LLMs: 18.7\% higher relevance in manual evaluation, 9.2\% increase in click-through rates, and 63\% faster inference speed. These advancements bridge the gap between academic research and industrial deployment, offering a robust solution for real-time conversational systems that balance latency constraints with user-centric guidance quality. 
%\end{abstract}

\begin{abstract}
The evolution of Large Language Models (LLMs) has significantly advanced multi-turn conversation systems, emphasizing the need for proactive guidance to enhance users' interactions. However, these systems face challenges in dynamically adapting to shifts in users' goals and maintaining low latency for real-time interactions. In the Baidu Search AI assistant, an industrial-scale multi-turn search system, we propose a novel two-phase framework to provide proactive guidance. The first phase, Goal-adaptive Supervised Fine-Tuning (G-SFT), employs a goal adaptation agent that dynamically adapts to user goal shifts and provides goal-relevant contextual information. G-SFT also incorporates scalable knowledge transfer to distill insights from LLMs into a lightweight model for real-time interaction. The second phase, Click-oriented Reinforcement Learning (C-RL), adopts a generate-rank paradigm, systematically constructs preference pairs from user click signals, and proactively improves click-through rates through more engaging guidance. This dual-phase architecture achieves complementary objectives: G-SFT ensures accurate goal tracking, while C-RL optimizes interaction quality through click signal-driven reinforcement learning. Extensive experiments demonstrate that our framework achieves 86.10\% accuracy in offline evaluation (+23.95\% over baseline) and 25.28\% CTR in online deployment (149.06\% relative improvement), while reducing inference latency by 69.55\% through scalable knowledge distillation.
\end{abstract}

\section{Introduction}

\begin{figure}
  \includegraphics[width=0.5\textwidth]{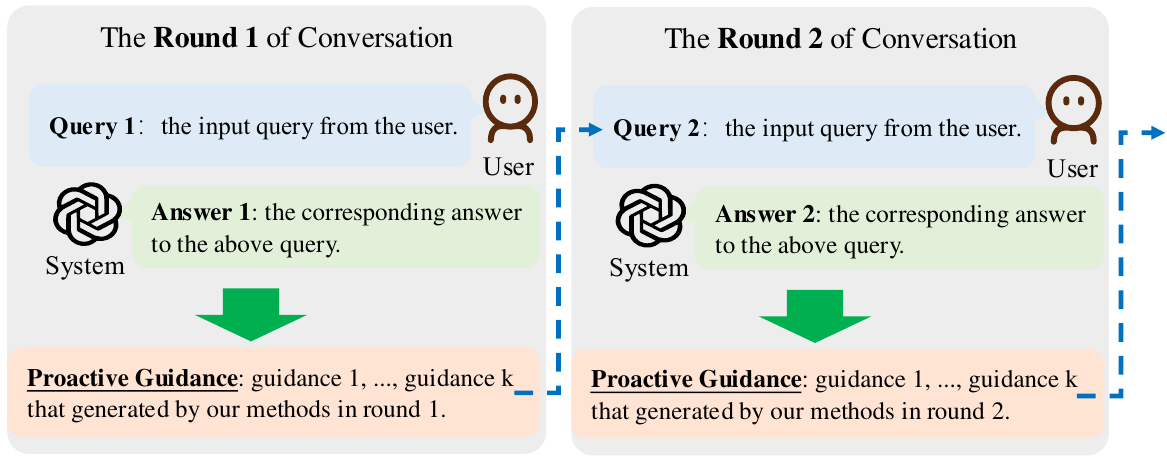}
  \caption{Illustration of the Proactive Guidance task in the multi-turn conversation system scenario. In each turn, given the user's query and the corresponding answer, our method generates $k$ proactive guidance to guide the user to click for the next turn of the conversation.}
  \label{fig:task-definition}
\end{figure}

%LLM变强、multi turn sys进步
The remarkable progress in Large Language Models (LLMs) \cite{achiam2023gpt, yang2024qwen2, grattafiori2024llama, guo2025deepseek} has propelled conversational AI systems into a new era, where they are increasingly capable of understanding users' queries and providing precise answers. This advancement has spurred the development of multi-turn conversation systems \cite{aliannejadi2020harnessing, vadhavana2024conversational, yi2024survey, zhang2025survey}. 

%对话系统关注guide，guide很重要
Contemporary systems are increasingly valued for their ability to anticipate and guide conversational turns \cite{zhang2018towards, gao2021advances, fang2024multi}. Instead of requiring users to precisely formulate their next query or even fully understand their own needs, systems can provide proactive guidance as follow-up questions that align with users' conversational goals and significantly enhance the convenience of interactions by minimizing the cognitive load on users. 
%重要，但是实现起来存在的问题
Despite their importance, crafting proactive guidance still remains challenging, particularly in multi-turn conversation systems where users' goals may undergo multiple shifts during interactions \cite{deng2023unified, bordes2016learning}.
%Guidance needs to dynamically adapt to shifts while maintaining the low latency necessary for real-time interaction.

%传统方法的问题
Traditional methods that utilize LLMs with historical conversation as contextual information have shown impressive results in guidance quality \cite{li2024incorporating, duan2025guidellm, feng2023large}.
However, they face several challenges when deployed in real-world scenarios. 
Firstly, these methods often struggle to dynamically adapt to changes in user conversational goals \cite{li2024incorporating}, as incorporating the entire conversation history can inadvertently introduce irrelevant information, which may result in misaligned guidance (i.e., query shifts from food allergy to the stock market may cause LLMs to persistently recommend food safety, losing track of the user's new conversational goal).
Secondly, redundant historical context, especially lengthy answers, introduces computational overhead and increased latency, severely affecting real-time interactions \cite{lapov2024dynamic}.
Lastly, the high computational demands of LLMs further amplify these issues, hindering their practicality in generating rapid responses.

%我们提出解决方法
%We propose an innovative framework that combines Goal-adaptive Supervised Fine-Tuning (G-SFT) and Click-oriented Reinforcement Learning (C-RL) to address these challenges, as illustrated in Figure~\ref{fig:task-definition}. 

To address these challenges, we propose an innovative framework that combines Goal-adaptive Supervised Fine-Tuning (G-SFT) with Click-oriented Reinforcement Learning (C-RL) to solve the proactive guidance task, as illustrated in Figure~\ref{fig:task-definition}. 

%new G-SFT
In the G-SFT phase, our Goal Adaptation Agent (GAA) dynamically identifies and adapts to user goal shifts through three core outputs: explicit goal analysis, shift detection signals, and concise goal-relevant summary. By replacing redundant historical context with these signals in the generation of guidance, we achieve 65.5\% faster processing in later turns and 10.18\% higher click-through rates.
Alongside this, scalable knowledge transfer distills LLMs' vast world knowledge into a more compact model, the G-SFT model, maintaining guidance quality while further reducing inference latency.

%new C-LR
The C-RL phase further optimizes the G-SFT model, leveraging user click signals to construct preference pairs for alignment. Various forms of reinforcement learning \cite{kaelbling1996reinforcement, schulman2017proximal, rafailov2023direct, amini2024direct, ethayarajh2024kto} have been proposed and implemented in conversation systems due to their ability to adapt responses to better align with user preferences. The key challenge lies in generating meaningful training samples of $k$ guidance from single-clicked guidance, as the model must provide $k$ guidance options per turn. We address this using a generate-rank paradigm: (1) training an augmentation model on $1$-pair click data, (2) generating diverse candidate guidance groups using Diverse Beam Search (DBS) \cite{vijayakumar2016diverse}, and (3) ranking and sampling $k$-pair data using a click estimator and a novel diversity-aware group sampling strategy. Experimental results demonstrate significant improvements, with accuracy increasing by 3.47\% and click-through rates increasing from 20.81\% to 25.28\% in industrial deployment environments.

Our contributions can be summarized as follows:
\begin{itemize}
    \item We introduce a goal adaptation agent that dynamically identifies and adapts to shifts in user goals, generating concise, goal-aligned summaries that streamline context for guidance generation without additional latency.
    \item We develop a generate-rank paradigm that leverages the DBS-based generation method, coupled with a group sampling strategy, to address the gap between single-preference data and multi-output requirements, thereby further enhancing the guidance quality.
    \item Comprehensive experiments demonstrate significant improvements in accuracy, task-related gains ($\Delta$GSB), and click-through rate, validating the effectiveness of our framework in real-world conversational search scenarios.
\end{itemize}

\section{Methodology}

\begin{figure*}[ht]
    \centering
    \includegraphics[width=\textwidth]{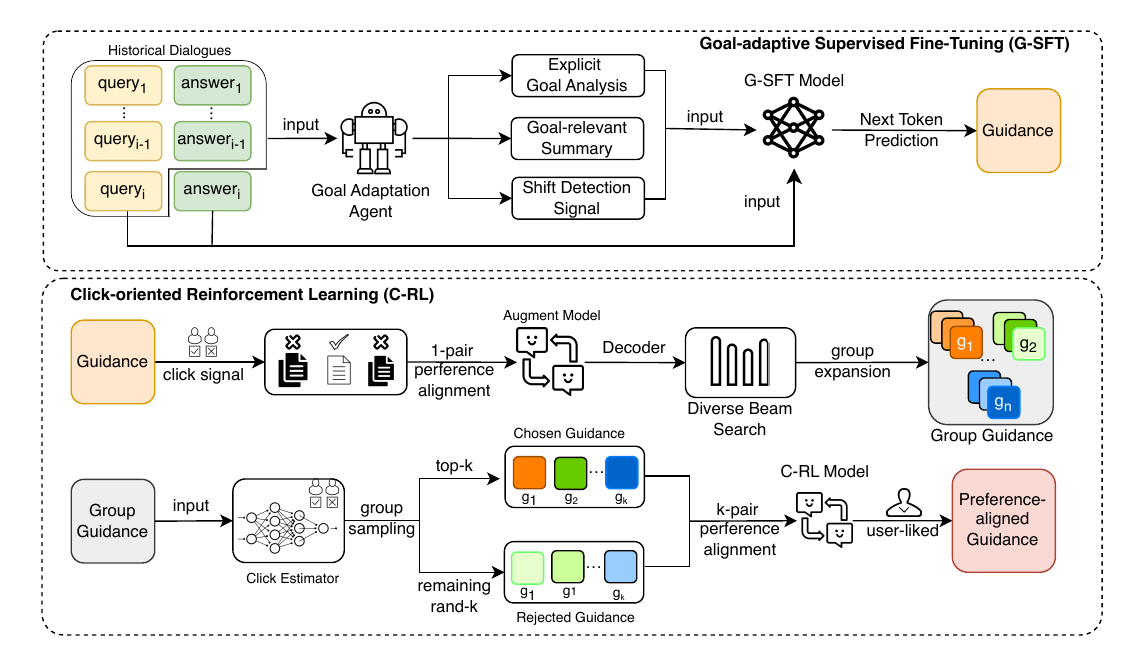}
    \caption{Architecture of the proposed framework.}
    \label{fig:architecture}
%%\textbf{\vspace{-0.5cm}}
\end{figure*}

In this section, we first provide a formal definition of the proactive guidance task in the multi-turn conversation system, then present our innovative two-phase framework, as illustrated in Figure \ref{fig:architecture}.

\subsection{Proactive Guidance}
\label{sec:PG}
The task aims to generate a set of guidance phrases, $G_i = \{G_{i1}, G_{i2}, \dots, G_{ik}\}$, during the $i$-th round of the conversation, where $k$ is a predefined constant. Specifically, in each round $i$, given user's query $Q_i$, the corresponding answer $A_i$ and contextual information $C_i$, our objective is to determine the optimal function $f_i^{*}$ to generate $G_i$ that maximizes the well-designed evaluation function $\mathbb{Y}$:
\begin{equation}\label{equ:pg-task}
f_i^{*}(Q_i, A_i, C_i) = \arg\max_{G_i}\mathbb{Y}(G_i \mid Q_i, A_i, C_i),
\end{equation}
where $\mathbb{Y}$ comprises two components: the offline and online evaluations. 
Offline evaluation, $\mathbb{Y}_{\text{offline}}$, assesses 1) Relevance: this evaluates the relevance of $G_i$ in the context of the conversation; 2) Applicability: this dimension measures the practical utility of $G_i$; 3) Diversity: this criterion evaluates the variety and breadth of $G_i$, ensuring a relatively comprehensive range of perspectives.
The $\mathbb{Y}_{\text{offline}}$ is conducted through manual scoring by trained annotators, with full evaluation criteria provided in Appendix \ref{appendix:eval}.
%\begin{equation}\label{equ:prior}
%    \mathbb{Y}_{\text{offline}} = \text{score}(G_i \mid Q_i, A_i, G_i).
%\end{equation}
Online evaluation, $\mathbb{Y}_{\text{online}}$, evaluates the effectiveness of the guidance $G_i$ in stimulating user engagement and promoting users' further interactions, which is quantified using the Click-Through Rate (CTR) metric.
%\begin{equation}\label{equ:posterior}
%    \mathbb{Y}_{\text{online}} = \text{CTR}(G_i).
%\end{equation}

\subsection{Goal-adaptive Supervised Fine-Tuning}
This phase is meticulously designed to produce a model capable of dynamically adapting to shifts in users' goals, providing high-quality guidance, and meeting the stringent latency requirements of industrial applications. 

\subsubsection{Goal Adaptation Agent}
Users' goals are defined as their explicit or implicit query intentions, which may undergo multiple shifts during interaction. By providing \textbf{E}xplicit goal analysis  $E_i$, goal-relevant \textbf{S}ummary $S_i$ and shift \textbf{D}etection signal $D_i$, all together as contextual information $C_i$, the Goal Adaptation Agent (GAA) effectively assists the guidance model in dynamically adapting to these shifts. 

The process is described in the following. In the initial round ($i=1$), the GAA is not activated. During the second round ($i=2$), it analyzes the current query $Q_2$ with the previous dialogue $(Q_1, A_1)$ to generate \{$E_i$, $S_i$, $D_i$\}. 
In subsequent rounds ($i > 2$), besides previous QA pair, the GAA additionally incorporates $S_{i-1}$ to seamlessly maintain context. 
This process is facilitated through the use of prompts, as described in Appendix \ref{appendix:GAA prompt}, which details the specific prompts employed by GAA.

\textbf{Explicit Goal Analysis.} GAA initially performs a detailed goal analysis by examining the correlation between the current query and the previous dialogue, identifying shifts and evolutions in the user’s goals; then it provides explicit textual descriptions of the current intentions and infers potential underlying needs.

\textbf{Goal-relevant Summary.} GAA generates concise, goal-aligned contextual information based on $\text{E}_i$ by (1) filtering goal-relevant segments from $A_{i-1}$ and $S_{i-1}$, and (2) inheriting pertinent information from $S_{i-1}$ while summarizing key points from $A_{i-1}$, omitting irrelevant details, to produce $S_i$, which focuses on the most relevant information, enabling the guidance agent to maintain coherence during dynamic goal shifts.

\textbf{Shift Detection Signal.} The detection signal $D_i$ serves as an indicator of whether a goal shift has occurred. When a goal shift is detected, $D_i$ prompts the system to reset $S_i$, thereby eliminating outdated information.% does not influence future interactions. 

%GAA结论
Two critical aspects of the GAA should be highlighted: First, the current answer $A_i$ is not used in GAA since it does not reflect the user's intent, allowing GAA to function simultaneously with answer generation and avoiding extra latency. Second, the contextual information $C_i$ provided by GAA is more concise than the raw chat history, significantly reduces the computational load for guidance generation, and ultimately decreases response latency.

\subsubsection{Scalable Knowledge Transfer}\label{sec:SKT}

% 解决的是部署问题
Although LLMs deliver impressive results, their latency can be prohibitive. Conversely, smaller models often lack the world knowledge needed to handle the diverse scenarios in reality. To address this, we propose a scalable knowledge transfer method.

Initially, we utilize LLMs to process various conversations, denoted as $Q_i$, $A_i$ and $C_i$, where $C_i$ is provided by GAA. Then LLMs are prompted to produce a chain of thought, $CoT_{i}$, paired with a list of $n$ guidance candidates, denoted as:
\begin{equation}
\{CoT_{i}, G_{i1}, \dots, G_{in}\} = \text{LLM}(Q_i,A_i,C_i).
\end{equation}
% 为了对齐先验评估标准
Subsequently, these $n$ candidates undergo a manual selection process based on $\mathbb{Y}_{\text{offline}}$, and $CoT_{i}$ is strategically discarded for efficiency, resulting in a refined subset of $k$ guidance, where $k < n$. We then fine-tune a significantly smaller model on this refined dataset through a loss function defined as follows:
\begin{equation}
L = -\sum_{t=1}^{T} \log P(y_t \mid y_{<t}, x),
\end{equation}
where $T$ is the length of the target sequence; $y_t$ is the target word at time step $t$; $y_{<t}$ denotes the sequence of words generated before time step $t$; $x$ is the input context.

Through scalable knowledge transfer, we have effectively equipped a more compact model, referred to as the G-SFT model, with the capability to offer insightful guidance whose quality rivals that of its larger counterparts. 

\subsection{Click-oriented Reinforcement Learning}
\label{sec:c-rl}
During the deployment of the G-SFT model, we collected substantial data on users’ interactions that inherently reflect user preferences. To fully exploit these valuable data, we introduced an innovative generate-rank paradigm, which effectively bridges the gap between the actual single-clicked guidance and the practical need for $k$ instances.

\subsubsection{Generate}
\label{sec:generate}
In this section, we demonstrate the process of generating multiple guidance phrases as candidates.

\textbf{Preference-Aligned Augmentation Model.} We leverage user interaction data to create training samples consisting of preference pairs. Each instance is composed of a question, an answer, and contextual information, collectively referred to as input $x$. The guidance clicked by a user is considered as the preferred response $y_w$, while the others are treated as dispreferred $y_l$, forming preference pairs $(x, y_w, y_l)$.
Then, we apply Direct Preference Optimization (DPO) \cite{rafailov2023direct} to the G-SFT model. The goal of the DPO loss function is to optimize the model's response probability, increasing the relative probability of the preferred response.
The formula is as follows:
\begin{align}
& \mathcal{L}_{\text{DPO}}(\pi_\theta; \pi_{\text{ref}}) = -\mathbb{E}_{(x, y_w, y_l) \sim \mathcal{D}} \Big[ \nonumber \\ 
& \log \sigma \Big( \beta \log \frac{\pi_\theta(y_w \mid x)}{\pi_{\text{ref}}(y_w \mid x)} - \beta \log \frac{\pi_\theta(y_l \mid x)}{\pi_{\text{ref}}(y_l \mid x)} \Big) \Big].
\end{align}
        
Through this process, we produce a preference-aligned model that has the ability to generate guidance that users are more likely to click on.

\textbf{DBS-based Decoding.} To generate multiple guidance outputs using the aligned model trained with single guidance, we incorporate the Diverse Beam Search (DBS) \cite{vijayakumar2016diverse} decoding strategy. DBS is an enhanced version of the beam search algorithm. It employs a grouping strategy that divides beams into multiple groups $\mathbf{Y}$ to explore different sequences independently. Additionally, DBS imposes a similarity penalty, discouraging the selection of tokens similar to those in other sequences. 

For a sequence $\mathbf{y}_{[t]}$, its dissimilarity against the group $g$ at time step $t$, $\mathbf{Y}^g_{[t]}$ , is measured as:
\begin{equation}
\Delta(\mathbf{y}_{[t]}, \mathbf{Y}^g_{[t]}) = \sum_{b=1}^{B'} \delta(\mathbf{y}_{[t]}, \mathbf{y}^g_{b,[t]}),
\end{equation}
where $\delta(\cdot, \cdot)$ quantifies sequence dissimilarity, e.g., a negative cost for each co-occurring n-gram in two sentences, distance between distributed sentence representations.

DBS decoding allows the aligned model to produce multiple responses in a single inference, providing guidance with significant differences in semantics, styles, or structures as candidates.

\subsubsection{Rank}\label{sec:rank}
This section describes how to construct preference pairs with $k$ guidance phrases.

\textbf{Click Estimator.} The Click Estimator is developed to predict the clicking likelihood of the guidance. It employs a sophisticated 12-layer ERNIE encoder \cite{sun2020ernie} that processes user interactions through a triplet format $(Q_i, G_{ij}, y)$, $j = 1,\dots,k$ and distinguishes between clicked ($y = 1$) and unclicked ($y = 0$) guidance. The training objective is:
\begin{align}
    \mathcal{L}(y, \hat{y}) = -\frac{1}{N} & \sum_{m=1}^{N} \Big[ \, y_m \cdot \log(\hat{y}_m) \nonumber \\
    & + (1 - y_m) \cdot \log(1 - \hat{y}_m) \Big],
\end{align}
where $\hat{y}$ denotes the predicted probability. This approach enables the click estimator to effectively predict the probability that a guidance $G_{ij}$ is clicked.

\textbf{Diversity-Aware Group Sample Strategy.} The sampling strategy that relies solely on click probability suffers from semantic redundancy, since the click estimator tends to assign similar scores to semantically equivalent guidance. 

Based on the traits of DBS, we propose a diversity-aware group sampling strategy that ensures semantic richness. It works as follows: (1) Organize candidates into \textit{n} groups where each group $P_i$ contains the $i$-th candidate from each beam, then select the highest-CTR candidate per group to yield \textit{n} diverse choices as a candidate pool $P$; (2) Apply Maximum Marginal Relevance (MMR) \cite{guo2010probabilistic} with
\begin{equation}
    \arg\max_{g_i \in P} \Big[ \lambda \cdot \text{CE}(g_i) - (1-\lambda) \cdot \max_{g_j \in S} \text{sim}(g_i, g_j) \Big],
\end{equation}
where $P$ denotes the candidate pool and $S$ denotes the selected set, $\text{CE}(\cdot)$ is the click probability predicted by the click estimator. $\lambda$ is a trade-off parameter that balances click probability and semantic diversity, which is set to 0.5 in our implementation. 
The selecting procedure starts with the guidance clicked by real users as the initial point, then selects $k-1$ guidance from $P$. These $k$ guidance are combined and seen as the preferred response. 
Then we randomly sampled $k$ guidance from the unselected  ones as dispreferred,
%while the dispreferred is randomly sampled $k$ from the unselected terms, 
ensuring that the maximum $\text{CE}(\cdot)$ score of the dispreferred guidance is less than the minimum score of the preferred guidance. The formats of training data are detailed in Appendix~\ref{app:train}.

Through this meticulous process, we create the $k$-pair preference-aligned dataset. Subsequently, we employed DPO to optimize the G-SFT model, resulting in the development of our final model being perceptible to user click preferences, referred to as the C-LR model. This model has significantly improved CTR in real-world application scenarios.

\section{Experiments}
To validate the effectiveness of our proposed method, we conducted comprehensive offline evaluations and online experiments within the Baidu Search AI assistant.

\subsection{Experimental Setup}

\textbf{Datasets.}
We evaluate our models using QA pairs collected from the Baidu Search AI assistant, an industrial-scale multi-round conversation system, to ensure authenticity and diversity. 
For the G-SFT model, we constructed a training set of 6,072 QA pairs following Section \ref{sec:SKT}. 
The C-RL model utilizes 12,000 preference pairs constructed using the generate-rank paradigm described in Section \ref{sec:c-rl}.

\textbf{Evaluation Metrics.} We evaluate the model’s performance using three metrics: 1) Accuracy (ACC): The proportion of guidance that meets the $\mathbb{Y}_{\text{offline}}$ as introduced in Section \ref{sec:PG}; 2) Good vs. Same vs. Bad ($\Delta$ GSB): Comparatively evaluates the performance of two models (details in Appendix~\ref{app:GSB}); 3) Click-Through Rate (CTR): The ratio of turns with click behavior to total turns.

\textbf{Baselines.} We adopt ERNIE Speed (21B) \cite{sun2020ernie, sun2021ernie}, a publicly accessible foundation model\footnote{https://cloud.baidu.com/product-s/qianfan\_home}, as our baseline model. The predefined number of guidance phrases $k$ is set to 3.
%with structured prompt, including task background, generation logic, guides the model to generate guidance based on the current conversational context. To fully preserve the model’s inherent knowledge inference capabilities, we employ a Zero-shot inference strategy.
%SKD-SFT leverages scalable knowledge transfer to distill the capabilities of guidance generation from GPT-4, with ERNIE-Speed-base serving as the student model.

\subsection{Implementation Details}

\textbf{G-SFT Phase.} We use ERNIE Speed as the base model, where the learning rate is 3e-6, the max sequence length is 4,096, the batch size is 16, and the model training epoch is 3. For scalable knowledge transfer, GPT-4o is chosen as the teacher model \cite{hurst2024gpt}.

\textbf{C-RL Phase.} Parameters are initialized with the best checkpoint of the G-SFT model. During the DPO process, the learning rate is set to 1e-6 with a batch size of 16, and the validation steps are set to 8. The training is conducted for 2 epochs. For DBS decoding parameters, the batch size is set to 16, the number of beam groups is 4, and the beam size within each group is 4.
%The parameters for the DBS blocking mechanism are set with BLOCK\_BS at 1.5 and BLOCK\_RATIO at 0.85. 

\subsection{Results and Analysis}

\begin{table}[t]
\caption{Performance comparison of different models.}
\label{tab:performance}
\centering
\begin{tabular}{l|>{\hspace{0cm}}c<{\hspace{0.1cm}}|>{\hspace{0cm}}c<{\hspace{0cm}}|>{\hspace{0cm}}c<{\hspace{0cm}}}
\toprule
\multirow{2}{*}{Model} & \multicolumn{2}{c|}{Offline} & \multicolumn{1}{c}{Online} \\ 
\cmidrule(lr){2-3}\cmidrule(lr){4-4}
 & \textbf{ACC} & $\Delta$ \textbf{GSB} & \textbf{CTR} \\ \midrule
BaseLine & 62.15\% & --- & 10.15\% \\ 
SKD model & 71.82\% & +2.43\% & 14.62\% \\ 
G-SFT model & 82.63\% & +4.24\% & 20.81\% \\ 
C-RL model & 86.10\% & +5.60\% & 25.28\% \\ 
\bottomrule
\end{tabular}
\begin{flushleft}
\small
\textit{Note:} \textbf{SKD model} refers to the model after Scalable Knowledge Transfer without the use of GAA. The \textbf{G-SFT model} is the model produced after the G-SFT stage of our proposed method, which incorporates both SKD and GAA. The\textbf{ C-RL model} is the G-SFT model fine-tuned with DPO on the dataset constructed using our proposed generate-rank method.
\end{flushleft}
\end{table}

\textbf{Overall Results.}
Experiments demonstrate significant improvements across offline and online metrics. As shown in Table~\ref{tab:performance}, the baseline model achieves 62.15\% ACC and 10.15\% CTR, while the C-RL model achieves improved performance with 86.10\% ACC, +5.60\% $\Delta$GSB and 25.28\% CTR. In particular, compared to the SKD model, the G-SFT model increases ACC by 10.81\% and CTR by 6.19\%, validating the superior goal management capabilities of GAA. Meanwhile, the C-RL phase further enhances CTR by 4.47\% with ACC gains (+3.47\%), demonstrating the ability of the C-RL model to capture implicit user preferences through click data. These results confirm the effectiveness of our two-phase framework, which excellently performs the task of proactive guidance. Appendix~\ref{app:case} provides a real sample.

\textbf{Consistency Analysis.} There is a strong correlation between offline and online metrics (Spearman's $\rho = 0.986$, $p < 0.01$), indicating that our proposed strategy not only improves objective accuracy but also effectively enhances user experience. The scalable knowledge transfer model shows improvements in ACC and CTR of +9.67\% and +4.47\% respectively, GAA with improvements of +10.18\%/+6.19\%, and C-RL with improvements of +3.47\%/+4.47\%. In particular, the excess gain in CTR of the reinforcement learning phase highlights its ability to capture implicit features of user goals through click behavior.

\textbf{Latency Analysis.} 
Our system achieves industrial-grade efficiency through two techniques: (1) Scalable knowledge transfer, transferring LLMs’ world knowledge to a more compact model and further removing the CoT, significantly reduces inference latency by 69.55\% (from 2.89s to 0.88s).
(2) by replacing raw chat history with GAA-generated concise contextual information, latency decreases by 65.5\% (3.25s → 1.12s). The combined optimizations enable real-time responsiveness with end-to-end latency around 1s, meeting industrial deployment requirements.

%Through the two techniques, our system achieves industrial-grade efficiency: (1) scalable knowledge transfer achieves a remarkable reduction  latency by 69.55\% (2.89s → 0.88s) via CoT elimination and compact model utilization; (2) by replacing raw chat history with GAA-generated concise contextual information, latency decreases by 65.5\% (3.25s → 1.12s). The combined optimizations enable real-time responsiveness with end-to-end latency around 1s, meeting industrial deployment requirements.

\subsection{Ablation Studies}

\begin{table}[t]
\caption{Ablation Studies of Goal Adaptation Agent (\textbf{GAA}).}
\label{tab:GAA}
\centering
\begin{tabular}{l|>{\hspace{0.1cm}}c<{\hspace{0.1cm}}|>{\hspace{0.1cm}}c<{\hspace{0.1cm}}|>{\hspace{0.1cm}}c<{\hspace{0.1cm}}}
\toprule
\multirow{2}{*}{Model} & \multicolumn{2}{c|}{Offline} & \multicolumn{1}{c}{Online} \\ 
\cmidrule(lr){2-3}\cmidrule(lr){4-4}
 & \textbf{ACC} & \boldmath{$\Delta$ \textbf{GSB}} & \textbf{CTR} \\ \midrule
SKD model & 71.82\% & --- & 14.62\% \\ 
+ S & 75.62\% & +2.97\% & 16.43\% \\ 
+ SD& 78.21\% & +3.11\% & 17.81\% \\ 
+ DE & 81.16\% & +3.67\% & 19.72\% \\ 
+ GAA & 82.63\% & +4.24\% & 20.81\% \\ 
\bottomrule
\end{tabular}
\begin{flushleft}
\small
\textit{Note:} \textbf{SKD model} refers to the model after Scalable Knowledge Transfer without the use of GAA. The table illustrates the impact of different components on model performance.\textbf{ S} represents the goal-relevant summary, \textbf{D} denotes the detection signal of goal shift, and \textbf{E} stands for explicit goal analysis.
\end{flushleft}
\end{table}

\textbf{Goal Adaptation Agent.} The ablation studies of the GAA in Table~\ref{tab:GAA} highlight the critical roles of its components: (1) goal-relevant $\textbf{S}$ummary, (2) $\textbf{D}$etection signal of goal shift, and (3) $\textbf{E}$xplicit goal analysis. The complete GAA achieves optimal performance with 82.63\% ACC and 20.81\% CTR, underscoring the importance of component synergy for effective multi-turn guidance.

Retaining only \textbf{S} results in a notable performance decrease (ACC -7.01\% , CTR -4.38\%), emphasizing the necessity of comprehensive goal management to maintain conversational coherence. Adding \textbf{D} helps recover some performance (ACC 78.21\%, CTR 17.81\%) by detecting goal shifts and prompting adjustments. However, \textbf{E} has a greater impact, achieving 81.16\% ACC and 19.72\% CTR, by providing a deeper understanding of user intentions. 
The results indicate that \textbf{D} and \textbf{E} are essential for maintaining coherent and context-aware guidance in multi-turn conversation.

\textbf{DBS Decoding Strategies.} This study examines the impact of the BEAM\_GROUP\_NUM \textbf{B} on generation quality using the DBS decoding strategy. As shown in Table~\ref{tab:RL}, setting \textbf{B} to 4 achieves the optimal balance with an ACC of 86.10\% and a CTR of 25.28\%. A group count of 1 limits the diversity, reducing CTR to 22.54\%, while 8 groups introduce noise, lowering CTR to 24.16\%. Notably, setting \textbf{B} to 2 maintains a high CTR of 24.78\% and improves decoding efficiency, offering a practical strategy for real-world deployment.

\begin{table}[t]
\caption{Ablation studies of DBS decoding parameters.}
\label{tab:RL}
\centering
\begin{tabular}{l|c|c|c}
\toprule
\multirow{2}{*}{Model} & \multicolumn{2}{c|}{Offline} & \multicolumn{1}{c}{Online} \\
\cmidrule(lr){2-3}\cmidrule(lr){4-4}
& \textbf{ACC} & \boldmath{$\Delta$ \textbf{GSB}} & \textbf{CTR} \\ \midrule
G-SFT model & 82.60\% & — & 20.81\% \\
$B = 1$ & 82.14\% & +2.88\% & 22.54\% \\
$B = 2$ & 84.87\% & +3.23\% & 24.78\% \\
$B = 4$ & 86.10\% & +3.60\% & 25.28\% \\
$B = 8$ & 84.31\% & +3.11\% & 24.16\% \\
\bottomrule
\end{tabular}
\begin{flushleft}
\small
\textit{Note:} \textbf{B} represents the BEAM\_GROUP\_NUM used in the diverse beam search decoding strategy. 
\end{flushleft}
\end{table}
\section{Conclusion}
In this paper, we propose a novel framework for proactive guidance in multi-turn conversation systems, integrating G-SFT with C-RL to address challenges in dynamic goal adaptation and real-time responsiveness. Our approach demonstrates significant improvements in both guidance quality and system efficiency.
Experimental results demonstrate that the framework effectively encourages user interaction and significantly increases click-through rates, highlighting its practical value in industrial scenarios.
\section{Future Work}

Despite the progress made in the proactive guidance for multi-turn conversation systems, there remain several areas for improvement and further investigation: 
\begin{itemize}
    \item \textbf{Refinement of summary reset mechanisms:} our current methodology resets $S_i$ when goal shifts are detected, failing to accommodate temporary shifts in user goals, resulting in loss of information when users return to previous intentions.
    Future enhancements could utilize a more sophisticated state-tracking system, allowing for a more flexible and coherent interaction experience.
    \item \textbf{Exploring more diverse baseline models:} 
    The comparison with baseline models in the current study has provided a foundational understanding of our framework’s capabilities. However, the rapid advancement in neural network architectures and language models suggests that integrating and comparing newer models could yield further insights.
    \item \textbf{Expansion of evaluation metrics:} the offline evaluation metrics used in this study, while comprehensive, could be expanded to include more diverse criteria that capture other aspects of user experience, such as user satisfaction or the system’s ability to handle unexpected queries. Future studies could explore additional metrics that provide a deeper understanding of the qualitative aspects of conversation.
\end{itemize}

By addressing these future directions, we aim to enhance the functionality and applicability of proactive guidance, paving the way for more intelligent, adaptable, and user-centric conversational agents. This continued research could have a profound impact on the development of AI-driven communication tools across various domains.

\bibliography{reference}

\clearpage
\appendix
\section{Prompt for the Goal Adaptation Agent}
\label{appendix:GAA prompt}
\begin{flushleft}
\texorpdfstring{\smash{\makebox[\linewidth][l]{\normalsize This appendix presents the structured prompt for the goal adaptation agent.}}}{}
\end{flushleft}
\begin{tcolorbox}[
    enhanced,
    colframe=lightred,
    colback=custombg,
    boxrule=1.0pt,
    arc=3pt,
    auto outer arc,
    fontupper=\sffamily,
    drop shadow,
    width=\textwidth,
    left=7pt,
    right=7pt,
    top=5pt,
    bottom=5pt
]
\textbf{Prompt:}
You are a \textbf{Goal-Tracking Model} specifically designed for multi-turn dialogue scenarios. Your task is to understand and track the user's evolving goals throughout the dialogue and produce coherent summaries that capture the history and progression of the conversation. This process involves preserving contextual continuity and relevance to the user's current objectives. To accomplish this, you will utilize the following inputs:

\begin{itemize}[leftmargin=1em,nosep]
    \item \textbf{[\colorbox{red}{$Q_i$}]}: The current user question in the dialogue, which may indicate a continuation of previous goals or the introduction of new goals.
    \item \textbf{[\colorbox{red}{$(Q_{i-1}, A_{i-1})$}]}: The immediate previous question and answer pair, providing context for $Q_i$ and potentially containing clues about changes in the user's intent since the last turn.
    \item \textbf{[\colorbox{red}{$S_{i-1}$}]}: A comprehensive summary of the dialogue history up to the interaction immediately preceding $Q_i$, encapsulating key points and actions taken that are relevant to the evolving goals of the user.
\end{itemize} 

\vspace{0.5\baselineskip}
\textbf{Task:}

\begin{enumerate}[label=(\arabic*),leftmargin=2em,align=left]
    \item \textbf{Explicit Goal Analysis:}
    \begin{itemize}[leftmargin=1em,nosep]
        \item Perform a detailed analysis of \textbf{[\colorbox{red}{$Q_i$}]} in the context of \textbf{[\colorbox{red}{$(Q_{i-1}, A_{i-1})$}]}, to detect nuanced changes in the user's goals. Provide a clear and explicit textual explanation that articulates the current user's intent, and infer any underlying or potential needs that may be driving this intent.
    \end{itemize}

    \item \textbf{Goal-relevant Summary:}
    \begin{itemize}[leftmargin=1em,nosep]
        \item Based on the results of the explicit goal analysis, selectively extract content from \textbf{[\colorbox{red}{$S_{i-1}$}]} and \textbf{[\colorbox{red}{$(Q_{i-1}, A_{i-1})$}]}, that is directly related to the user's current goals. Integrate these key points into a new, updated summary \textbf{[\colorbox{red}{$S_i$}]}, ensuring that it is concise yet comprehensive. Prune any elements that are no longer relevant to the current context or the user's goals to maintain focus and clarity in the evolving conversation.
    \end{itemize}

    \item \textbf{Detection Signal:}
    \begin{itemize}[leftmargin=1em,nosep]
        \item Provide a detection signal \textbf{[\colorbox{red}{$D_i$}]} that indicates whether a goal transition has occurred between the previous turn and the current turn. If such a transition is detected, trigger a reset of \textbf{[\colorbox{red}{$S_i$}]} to ensure that the summary remains relevant and does not retain outdated information that could interfere with the user's current goal orientation.
    \end{itemize}
\end{enumerate}

\textbf{Expected Output Format:}
\vspace{0.5\baselineskip}

The expected output should be a structured JSON object, as follows:
\begin{verbatim}
{
  "explicitGoalAnalysis": "Description of the user's current intent, and inferred
  potential needs of the user",
  "goalRelevantSummary": "Coherent summary incorporating key points relevant
  to the user's current goals",
  "detectionSignal": "Boolean indicating whether a goal transition has been
  detected"
}
\end{verbatim}

\end{tcolorbox}

\clearpage
\section{Data format of G-SFT and C-RL}
\label{app:train}

\subsection{Prompt format}
\begin{flushleft}
\texorpdfstring{\smash{\makebox[\linewidth][l]{\normalsize Here is the detailed prompt used for G-SFT and C-RL.}}}{}
\end{flushleft}
\begin{tcolorbox}[
    enhanced,
    colframe=lightred,
    colback=custombg,
    boxrule=1.0pt,
    arc=3pt,
    auto outer arc,
    fontupper=\sffamily,
    drop shadow,
    width=\textwidth,
    left=7pt,
    right=7pt,
    top=5pt,
    bottom=5pt
]
\textbf{Background:}
As a Proactive Guidance Model, you are tasked with enhancing user experience in a multi-turn dialogue system by predicting potential future inquiries. Through careful analysis of the current and past interactions, you will help drive the conversation towards fulfilling the user's objectives.

\textbf{Input Explanation:}
The following elements are provided for your analysis:
\begin{itemize}
    \item Current round's user query (\textbf{[\colorbox{red}{$Q$}]}).
    \item The corresponding system's answer (\textbf{[\colorbox{red}{$A$}]}).
    \item Contextual information from previous rounds, which includes:
        \begin{itemize}
            \item A summary of the dialogue thus far (\textbf{[\colorbox{red}{$S$}]}).
            \item Explicit goal analysis, detailing the objectives and needs of the user (\textbf{[\colorbox{red}{$E$}]}).
        \end{itemize}
\end{itemize}

\textbf{Thought Process:}
In predicting the user's next questions, you should:
\begin{enumerate}
    \item Assess if the current round’s answer (\textbf{[\colorbox{red}{$A_n$}]}) has adequately addressed the user's query (\textbf{[\colorbox{red}{$Q_n$}]}).
    \item Utilize the contextual information, particularly the summary and explicit goal analysis, to comprehend the user’s continuous journey and objectives within the dialogue.
    \item Anticipate the user's potential next steps by considering the dialogue's progression and any identified goals or needs.
    \item Generate $k$ relevant and contextually appropriate questions as guidance that the user might ask next.
\end{enumerate}

\textbf{Output Format Requirements:}
Present your predictions structured as follows:

\begin{verbatim}
Guidan_1\n...\nGuidance_k
\end{verbatim}
\end{tcolorbox}

\subsection{Response format:}
\begin{flushleft}
\texorpdfstring{\smash{\makebox[\linewidth][l]{\normalsize Here shows the response format of different tasks.}}}{}
\end{flushleft}
\begin{tcolorbox}[
    enhanced,
    colframe=lightred,
    colback=custombg,
    boxrule=1.0pt,
    arc=3pt,
    auto outer arc,
    fontupper=\sffamily,
    drop shadow,
    width=\textwidth,
    left=7pt,
    right=7pt,
    top=5pt,
    bottom=5pt
]
\textbf{For G-SFT:}
\begin{verbatim}
response: Guidan_1\nGuidance_2\nGuidance_3
\end{verbatim}

\textbf{For 1-pair DPO(Augmentation model as in section~\ref{sec:generate}):}
\begin{verbatim}
Chosen: Guidance(clicked)
Rejected: Guidance(unclicked)
\end{verbatim}

\textbf{For k-pair DPO(C-RL model as in section~\ref{sec:c-rl}):}
\begin{verbatim}
Chosen: Guidance_pos1\nGuidance_pos2\nGuidance_pos3
Rejected: Guidance_neg1\nGuidance_neg2\nGuidance_neg3
\end{verbatim}
note: $Guidance\_pos*$ stands for the chosen guidance sampled through the method in section~\ref{sec:rank}, while $Guidance\_neg*$ stands for rejected guidance.
\end{tcolorbox}

\clearpage
\section{Showcase}
\label{app:case}
\begin{figure}
  \includegraphics[width=0.5\textwidth]{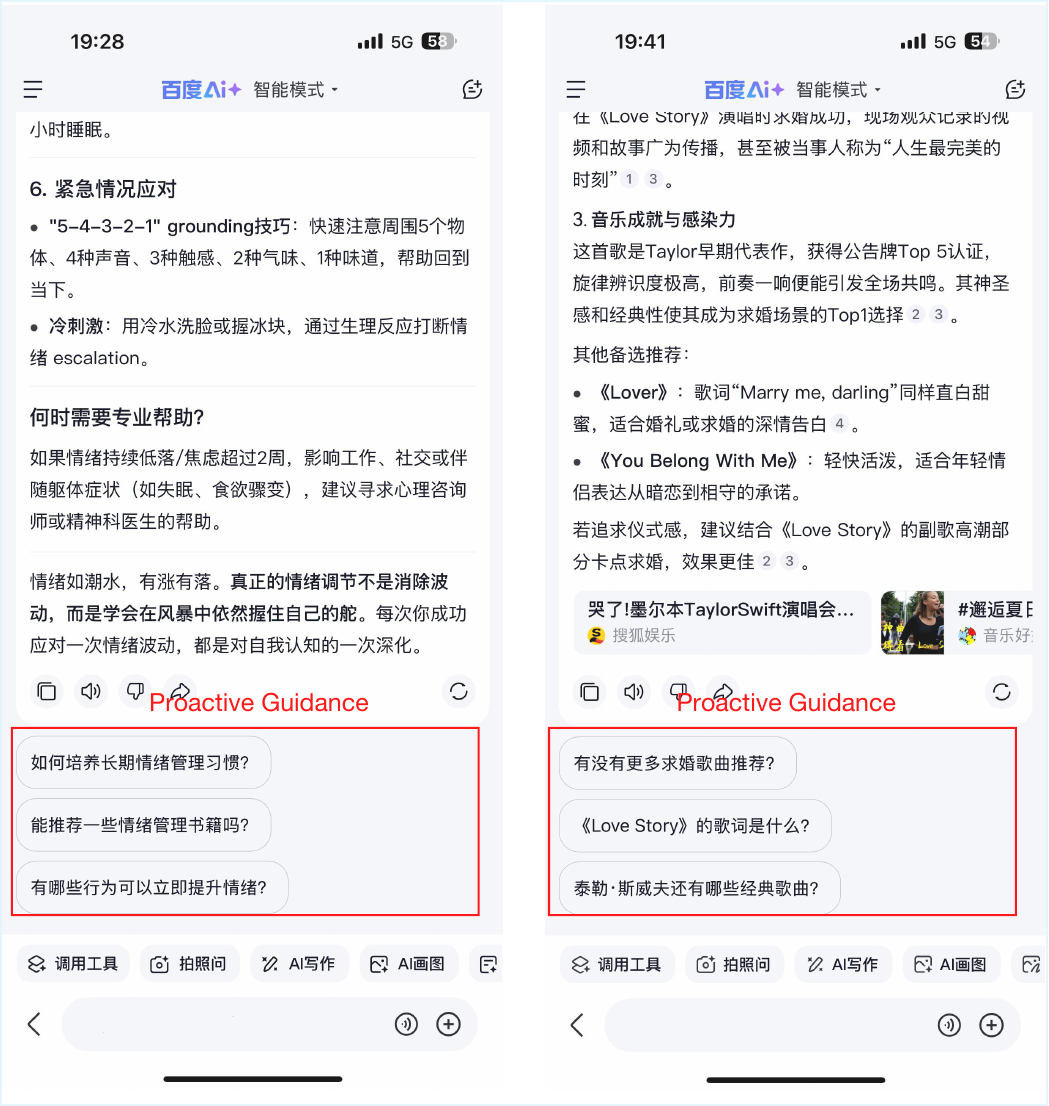}
  \caption{Proactive guidance in Baidu Search AI assistant. The left query is "How to manage emotions?" and the right query is "Which Taylor Swift song is suitable for a marriage proposal?"}
  \label{fig:case}
\end{figure}

Figure \ref{fig:case} demonstrates proactive guidance in the Baidu Search AI assistant, an industrial-scale multi-turn conversation system.

On the left side of the image, the user poses the question "How to manage emotions?" The guidance is organized into three key areas: cultivating long-term emotional management habits, recommending books on emotional management, and identifying actions for immediate mood improvement. Cultivating long-term habits focuses on sustainable practices, building resilience over time. Book recommendations offer resources for deeper learning, while immediate mood improvement actions provide practical strategies for real-time relief. This structured approach effectively refines the inquiry into specific, actionable advice, enhancing user satisfaction.

On the right side of the image, the user inquires, "Which Taylor Swift song is suitable for a marriage proposal?" The guidance here is thoughtfully structured into three suggestions: Are there any more song recommendations for a proposal? What are the lyrics to "Love Story"? What other classic songs does Taylor Swift have? Each recommendation serves a distinct purpose, ensuring comprehensive support for the user's inquiry. The first expands song options, enhancing satisfaction by offering a wider array of choices. The second caters to users interested in song lyrics, allowing a deeper connection with the thematic elements. The third broadens the user's musical horizon with classic Taylor Swift songs, aiding in discovering songs that resonate with their proposal vision.

Overall, the guidance in both scenarios is diverse and non-overlapping, addressing potential user goals and enhancing engagement through structured, actionable advice.

\section{Evaluation Criteria}
\label{appendix:eval}

This appendix outlines the evaluation criteria used for assessing the effectiveness of the guidance phrases generated during the conversation rounds. Our evaluation framework consists of three main components: relevance, applicability, and diversity. Each component is crucial for ensuring the quality and utility of the guidance provided. The evaluation is conducted by trained annotators based on the following detailed criteria:

\subsection{Relevance}
\begin{itemize}
    \item \textbf{Contextual Relevance}: The guidance phrases must be directly related to the user's query and the ongoing conversation. They should address the user's needs without introducing unrelated topics.
    \item \textbf{Coherence}: The phrases should maintain logical consistency with the conversation history, avoiding contradictions and repetition.
\end{itemize}

\subsection{Applicability}
\begin{itemize}
    \item \textbf{Intent Clarification}: When the user's intent is unclear or comprises multiple potential directions, the guidance should help the user to clarify their intent.
    \item \textbf{Identifying Hidden Demands}: If the current query is only part of the user's fundamental needs, the guidance should aim to uncover underlying requirements, offering comprehensive or extended guidance.
    \item \textbf{Personalized Information Supplementation}: When the user's intent is clear but requires personalized information, the guidance should prompt the user to provide necessary context for a tailored response.
\end{itemize}

\subsection{Diversity}
\begin{itemize}
    \item \textbf{Comprehensiveness}: The guidance should cover a wide range of dimensions or options. It should be supported by expert knowledge or strong a posteriori information justifying the necessity of each guidance element.
    \item \textbf{Mutual Exclusivity}: The guidance should not repeat or overlap with the user's original query or with content already adequately addressed in previous answers. Different guidance options should be distinct from one another, avoiding intersections or inclusions.
\end{itemize}

\subsection{Redline Criteria}
\begin{itemize}
    \item \textbf{Legal and Ethical Compliance}: Guidance must not violate national laws, involve sensitive political or adult content, or touch on sensitive topics.
    \item \textbf{Accuracy and Truthfulness}: The information provided must be factual and free from rumors or misinformation.
    \item \textbf{Emotional Impact}: Guidance should avoid content that is excessively violent, discomforting, or sensationalist, such as exaggerated or eye-catching lowbrow titles.
\end{itemize}

%\subsection{Scoring Methodology}
%Each set of $k$ guidance phrases is evaluated as a whole using a binary scoring system:

%\begin{itemize}
%    \item \textbf{1 Point}: Awarded if all guidance phrases in the set fully meet the criteria.
%    \item \textbf{0 Points}: Given if any guidance phrase in the set fails to meet any of the criteria.
%\end{itemize}

%This holistic scoring approach ensures that the quality of each set is assessed comprehensively.

%\subsection{Accuracy Metric}
%The accuracy of the guidance generation is calculated as follows:

%$$ \text{Accuracy} = \frac{\text{Total points awarded across all sets}}{\text{Total number of sets}} $$

%This metric provides a quantitative measure of the model's effectiveness in generating guidance that meets the established criteria.

%The evaluation is conducted offline by a team of trained annotators who score the guidance phrases based on these criteria. The comprehensive nature of this evaluation process ensures that the guidance provided is not only relevant and applicable but also diverse and ethically sound.

\section{Good vs. Same vs. Bad (GSB) Calculation Details} \label{app:GSB}
Good vs. Same vs. Bad (GSB) is a metric judged by professionally trained annotators. For each user query, annotators are presented with the answer, historical conversations, and the guidance generated from both model A and model B. Based on the quality of the guidance, annotators independently assign one of the following labels:
\begin{itemize}
    \item \textbf{Good}: Results from model A are better than model B.
    \item \textbf{Bad}: Results from model B are better than model A.
    \item \textbf{Same}: Results from model A and model B are of equal quality (either good or bad).
\end{itemize}

To quantify the human evaluation, we use a unified metric $\Delta$GSB, defined as:
$$
    \Delta \text{GSB} = \frac{\# \text{Good} - \# \text{Bad}}{\# \text{Good} + \# \text{Same} + \# \text{Bad}}.
$$

\end{document}